\documentclass[runningheads]{llncs}

\usepackage{graphicx}
\usepackage{booktabs}
\usepackage{amsmath,amssymb}
\usepackage{url}
\usepackage{hyperref}

\title{Explainability-Inspired Layer-Wise Pruning of Deep Neural Networks for Efficient Object Detection}
\titlerunning{Explainability-Inspired Layer-Wise Pruning}

\author{Abhinav Shukla\inst{1} \and
Nachiket Tapas\inst{1}}
\authorrunning{A. Shukla et al.}
\institute{Chhattisgarh Swami Vivekanand Technical University, Bhilai, India 
}
\begin{document}
% ======================
\maketitle

% ======================
% Paper content (UNCHANGED)
% ======================
\begin{abstract}
Deep neural networks (DNNs) have achieved remarkable success in object detection tasks, but their increasing complexity poses significant challenges for deployment on resource-constrained platforms. 
While model compression techniques such as pruning have emerged as essential tools, traditional magnitude-based pruning methods do not necessarily align with the true functional contribution of network components to task-specific performance. 
In this work, we present an explainability-inspired, layer-wise pruning framework tailored for efficient object detection. 
Our approach leverages a SHAP-inspired gradient--activation attribution to estimate layer importance, providing a data-driven proxy for functional contribution rather than relying solely on static weight magnitudes. 
We conduct comprehensive experiments across diverse object detection architectures, including ResNet-50, MobileNetV2, ShuffleNetV2, Faster R-CNN, RetinaNet, and YOLOv8, evaluating performance on the Microsoft COCO 2017 validation set. 
The results show that the proposed attribution-inspired pruning consistently identifies different layers as least important compared to L1-norm-based methods, leading to improved accuracy--efficiency trade-offs. 
Notably, for ShuffleNetV2, our method yields a 10\% empirical increase in inference speed, whereas L1-pruning degrades performance by 13.7\%. 
For RetinaNet, the proposed approach preserves the baseline mAP (0.151) with negligible impact on inference speed, while L1-pruning incurs a 1.3\% mAP drop for a 6.2\% speed increase. 
These findings highlight the importance of data-driven layer importance assessment and demonstrate that explainability-inspired compression offers a principled direction for deploying deep neural networks on edge and resource-constrained platforms while preserving both performance and interpretability.
\end{abstract}

\section{Introduction}
\label{sec:introduction}

Deep neural networks (DNNs) have enabled substantial progress in computer vision, achieving strong performance across a wide range of tasks, including object detection in complex and dynamic environments. 
Their hierarchical representations allow models to capture rich semantic and spatial information, making them suitable for applications such as intelligent surveillance, robotics, autonomous driving, and mobile vision systems. 
However, the growing depth and complexity of modern detection architectures significantly increase computational cost, memory usage, and energy consumption, which limits their deployment on edge and resource-constrained platforms.

Model compression has therefore become a key enabler for practical deployment of DNNs. 
Techniques such as pruning, which removes less influential network components, and quantization, which reduces numerical precision, aim to reduce computational and memory overhead while preserving predictive performance~\cite{Deng2020ModelCompressionSurvey}. 
Among these, pruning is particularly attractive due to its flexibility and compatibility with existing architectures. 
Most pruning strategies rely on magnitude-based heuristics or statistical properties of weights~\cite{Molchanov2017PruningCNNs}. 
While effective in many cases, such criteria do not necessarily reflect the functional contribution of network components to the end task, especially in complex pipelines such as object detection.

Explainable Artificial Intelligence (XAI) provides tools for interpreting the internal behavior of deep models by attributing model predictions to individual components. 
Methods such as Layer-wise Relevance Propagation (LRP)~\cite{Montavon2019LRP}, DeepLIFT~\cite{Shrikumar2017DeepLIFT}, and Neuron Importance Score Propagation (NISP)~\cite{Yu2018NISP} produce data-dependent importance estimates for neurons, filters, or layers. 
This capability has inspired recent work that integrates attribution mechanisms into model compression pipelines, enabling pruning decisions to be guided by measures of functional relevance rather than static parameter statistics~\cite{Becking2022ECQ}. 
Such approaches suggest that attribution-informed criteria can offer improved robustness compared to purely magnitude-based heuristics.

Despite these advances, important gaps remain. 
Many explainability-inspired compression methods require additional supervision, carefully designed baselines, or task-specific tuning, which limits their scalability. 
Moreover, much of the existing literature focuses on image classification, with comparatively limited investigation of attribution-guided pruning in object detection, where architectural complexity, multi-scale feature processing, and efficiency constraints pose additional challenges. 
In this context, it remains unclear how attribution-inspired importance measures behave at the layer level and whether they can reliably guide structured pruning decisions in detection models.

In this work, we propose an explainability-inspired, layer-wise pruning framework for object detection networks. 
Rather than computing exact game-theoretic attributions, we adopt a gradient--activation-based attribution inspired by SHAP and DeepLIFT to estimate the functional contribution of network layers in a data-driven manner. 
These attribution scores are used to globally rank layers and deactivate those with minimal estimated contribution, enabling structured pruning without architectural rewiring. 
Through extensive experiments on diverse detection architectures and datasets, we demonstrate that the proposed approach identifies different pruning candidates than traditional magnitude-based methods and yields favorable accuracy--efficiency trade-offs. 
Our results suggest that attribution-inspired layer ranking offers a practical and interpretable alternative for guiding pruning decisions in complex vision systems.

The remainder of the paper is organized as follows. 
Section~\ref{sec:methodology} describes the proposed attribution-inspired pruning framework. 
Section~\ref{sec:Experiments} presents the experimental setup and evaluation results. 
Finally, Section~\ref{sec:conclusion} concludes the paper and discusses future directions.

\section{Related Work}
\label{sec:relatedwork}

The integration of explainability with model compression has emerged as a promising direction for improving both the efficiency and interpretability of deep neural networks. 
Early studies demonstrated that relevance-based attribution methods, such as Layer-wise Relevance Propagation (LRP), can provide meaningful estimates of component importance and can be leveraged to guide global pruning decisions~\cite{Yeom2021PruningByExplaining,Montavon2019LRP}. 
These works showed that parameters with low relevance scores could be removed with limited impact on predictive performance, motivating the use of explainability as a principled pruning signal.

Building on this idea, Becking \emph{et al.}~\cite{Becking2022ECQ} introduced explainability-inspired quantization for low-bit and sparse neural networks, demonstrating that relevance-aware compression can outperform magnitude-based strategies in preserving accuracy. 
Similarly, interpretability-driven filter and channel pruning methods have exploited attribution scores to guide structured model reduction~\cite{Yao2021InterpretabilityFilterPruning}. 
Collectively, these approaches suggest that attribution-informed criteria better capture functional importance than static weight statistics.

Several works have explored explainability-guided pruning in dynamic or online settings. 
Sabih \emph{et al.}~\cite{Sabih2022DyFiP} proposed DyFiP, which applies DeepLIFT-based relevance scores to dynamically prune filters during inference, achieving favorable efficiency–relevance trade-offs. 
Related efforts have incorporated explainability into ensemble pruning and online learning scenarios, demonstrating improved robustness under non-stationary data distributions~\cite{Saadallah2022ExplainableEnsemble}.

More recently, the relationship between pruning, representation quality, and interpretability has been examined in the context of vision models. 
Cassano \emph{et al.}~\cite{Cassano2024PruningVision} analyzed when pruning benefits vision representations, showing that moderate pruning can improve object discovery and explanation clarity, whereas excessive pruning degrades both accuracy and interpretability. 
Weber \emph{et al.}~\cite{Weber2023InfluencePruningExplainability} further quantified the impact of pruning on CNN explainability, highlighting the risk of interpretability loss under aggressive compression.

Other studies have focused on sparsifying explanations rather than modifying the underlying model. 
Sarmiento \emph{et al.}~\cite{Sarmiento2024SparseExplanations} proposed pruning strategies applied directly to relevance maps to obtain sparse, input-dependent explanations without altering the global network structure. 
In a broader perspective, Soroush \emph{et al.}~\cite{Soroush2025CompressingXAI} presented a unified framework for compressing deep neural networks using relevance scores for both pruning and mixed-precision quantization, achieving substantial reductions in model size without accuracy loss.

Despite these advances, most existing work focuses on image classification or pruning at the level of individual neurons, filters, or explanations. 
Comparatively little attention has been given to explainability-inspired, layer-wise pruning for object detection, where architectural complexity, multi-scale feature representations, and deployment constraints introduce additional challenges. 
In contrast to prior studies, our work investigates global, layer-wise attribution-inspired pruning applied directly to modern object detection architectures, providing an empirical analysis of accuracy–efficiency trade-offs in a detection-centric setting.

\begin{table*}[ht]
\centering
\caption{Comparison of explainability-inspired compression and pruning approaches.}
\scalebox{0.68}{
\begin{tabular}{|l|l|l|l|l|l|l|l|l|p{2cm}|}
\hline
\textbf{Work} & \textbf{XAI Method} & \textbf{Compression} & \textbf{Mode} & \textbf{Quant.} & \textbf{Target} & \textbf{Object} & \textbf{Layer-} & \textbf{Interp.–} & \textbf{Key} \\
 &  & \textbf{Target} &  &  & \textbf{Domain} & \textbf{Detection} & \textbf{wise} & \textbf{Efficiency} & \textbf{Contribution} \\
\hline
\cite{Yeom2021PruningByExplaining} & LRP & Model & Global & -- & Classification & -- & -- & Yes & LRP-guided global pruning \\

\cite{Becking2022ECQ} & LRP & Weight & Global & Yes & Classification & -- & -- & Yes & Explainability-inspired quantization \\

\cite{Yao2021InterpretabilityFilterPruning} & Various & Filter/Channel & Global & -- & Classification & -- & -- & Yes & Attribution-based filter pruning \\

\cite{Sabih2022DyFiP} & DeepLIFT & Filter & Dynamic & -- & CNNs & -- & -- & Yes & Online relevance-aware pruning \\

\cite{Cassano2024PruningVision} & Various & Model & Global & -- & Vision & Yes & -- & Yes & Effect of pruning on object discovery \\

\cite{Weber2023InfluencePruningExplainability} & LRP & Model & Global & -- & CNNs & -- & -- & Yes & Quantifies explainability loss \\

\cite{Sarmiento2024SparseExplanations} & LRP & Explanation & Local & -- & Classification & -- & -- & No & Sparse local explanations \\

\cite{Saadallah2022ExplainableEnsemble} & Saliency & Ensemble & Online & -- & Cls./Time Ser. & -- & -- & Yes & Explainable ensemble pruning \\

\cite{Soroush2025CompressingXAI} & LRP & Model & Global & Yes & Various & -- & Yes & Yes & Unified XAI-based compression \\

\hline
\textbf{This} & Gradient-based & Model & Global & -- & Object Detection & Yes & Yes & Yes & Layer-wise pruning for \\
\textbf{Work}& attribution (SHAP- & & & & & & & & modern detectors using \\
\textbf{} & inspired) & & & & & & & & attribution-inspired ranking \\

\hline
\end{tabular}
}
\label{tab:relatedwork_comparison}
\end{table*}

\section{Methodology}
\label{sec:methodology}

We propose an explainability-inspired framework for layer-wise pruning of object detection networks. 
The key idea is to estimate the functional contribution of each layer using a data-dependent attribution score and to use this estimate to guide structured pruning decisions. 
Our approach is evaluated through a controlled comparison against a standard magnitude-based pruning baseline. 
This section first introduces the baseline importance metric, then details the proposed attribution-inspired layer importance formulation, and finally describes the pruning procedure.

\subsection{L1-Norm Magnitude Pruning (Baseline)}
As a baseline, we adopt a commonly used structured pruning strategy based on the L1-norm of network weights. 
This method assumes that layers with smaller aggregate weight magnitudes contribute less to the model’s output and can therefore be pruned with minimal impact on performance. 
For each convolutional layer $l$ with weights $W_l$, an importance score $S_l^{\text{L1}}$ is computed as
\begin{equation}
    S_l^{\text{L1}} = \sum_{i} |W_{l,i}|,
    \label{eq:l1_norm}
\end{equation}
where $W_{l,i}$ denotes the individual weight parameters of layer $l$. 
This score provides a static, data-independent estimate of layer importance. 
While computationally efficient, it does not capture how the layer is utilized during inference on real data and serves primarily as a structural reference point for comparison.

\subsection{Attribution-Inspired Contribution Estimation (Proposed)}
To obtain a data-driven estimate of layer importance, we introduce an attribution-inspired contribution score based on gradient–activation interactions. 
We emphasize that our method does not compute exact SHAP values, which are derived from cooperative game theory and are computationally infeasible for layer-wise attribution in deep object detection models. 
Instead, we adopt a SHAP-inspired approximation that is closely related to gradient-based attribution methods such as GradientSHAP and DeepLIFT~\cite{lundberg2017unified,Shrikumar2017DeepLIFT}. 
The objective is not to approximate Shapley values, but to estimate the functional influence of each layer on the task loss in a computationally efficient manner.

For a given layer $l$, we define its contribution score $S_l^{\text{attr}}$ using the element-wise product of the layer’s output activations $A_l$ and the gradient of the loss $\mathcal{L}$ with respect to those activations. 
This gradient–activation product captures both the magnitude of a layer’s response and the sensitivity of the final objective to that response, providing a first-order approximation of contribution. 
The absolute value is taken to measure contribution magnitude irrespective of sign, since both positive and negative influences indicate functional relevance. 
Formally, the score is computed as
\begin{equation}
    S_l^{\text{attr}} = \mathbb{E}_{\mathbf{x} \sim D} \left[ \sum_{i} \left| \nabla_{A_{l,i}(\mathbf{x})} \mathcal{L} \cdot A_{l,i}(\mathbf{x}) \right| \right],
    \label{eq:shap_score}
\end{equation}
where the expectation is approximated over a representative mini-batch $\mathbf{x}$ sampled from the validation set $D$, and $A_{l,i}$ denotes the $i$-th activation element of layer $l$.

To compute these scores efficiently, we employ PyTorch forward and ba- ckward hooks to capture activation and gradient tensors during a single forward–backward pass. 
This avoids any modification to the underlying network architecture and introduces negligible computational overhead.

\begin{figure}[t]
    \centering
    \includegraphics[scale=0.45]{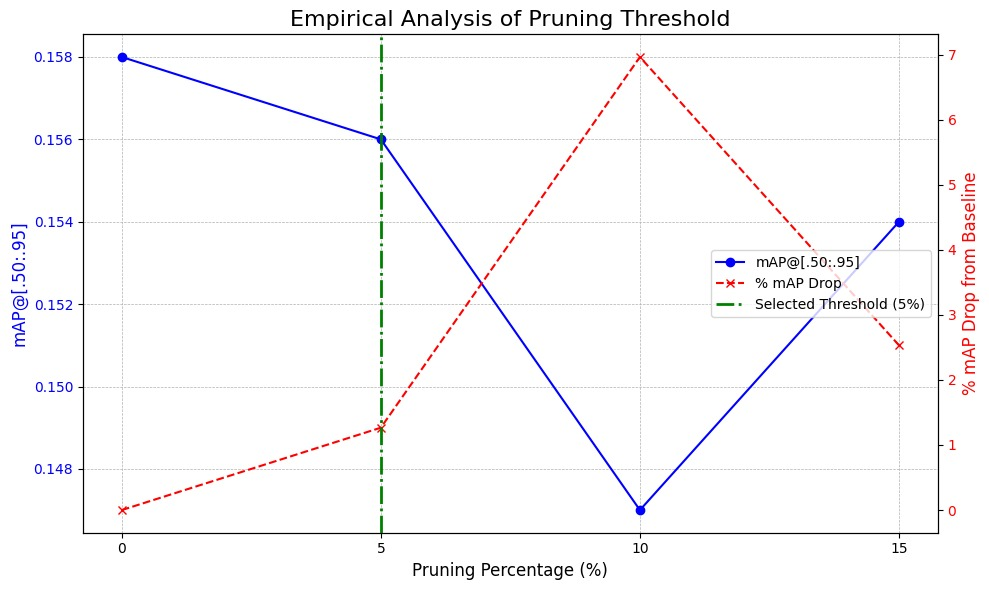}
    \caption{Empirical analysis of pruning thresholds illustrating the trade-off between detection accuracy (mAP drop) and pruning rate. A pruning rate of 5\% provides a favorable balance between efficiency and accuracy, while more aggressive pruning leads to disproportionate performance degradation.}
    \label{fig:pruning_threshold}
\end{figure}

\subsection{Layer Pruning Procedure}
The final step of the framework is layer-wise pruning based on the computed importance scores. 
Our goal is to reduce model complexity while preserving detection performance, without introducing architectural instability. 
To this end, we conduct a preliminary analysis to examine the impact of different pruning rates, as shown in Fig.~\ref{fig:pruning_threshold}. 
We observe that pruning rates of 10\% or higher lead to substantial accuracy degradation, whereas a pruning rate of 5\% achieves a meaningful efficiency gain with limited impact on performance.

Based on this analysis, we adopt a global pruning rate of 5\% for all experiments. 
For both the L1-norm baseline and the proposed attribution-inspired method, the pruning procedure consists of the following steps:
\begin{enumerate}
    \item \textbf{Score Computation:} Compute an importance score ($S_l^{\text{L1}}$ or $S_l^{\text{attr}}$) for all prunable convolutional layers.
    \item \textbf{Ranking:} Rank layers globally from least to most important according to their scores.
    \item \textbf{Pruning:} Deactivate the bottom 5\% of layers by zeroing their associated weights, effectively bypassing their contribution during inference while keeping the network graph unchanged.
\end{enumerate}
No post-pruning fine-tuning is performed, allowing us to isolate the direct effect of the pruning criteria on accuracy and efficiency. 
This consistent and attribution-aware procedure enables a controlled comparison between magnitude-based and explainability-inspired layer ranking strategies.

\section{Experiments}
\label{sec:Experiments}
To rigorously evaluate the proposed attribution-inspired layer-wise pruning framework, we conduct a comprehensive set of experiments on diverse object detection architectures. 
The evaluation is designed to assess the effectiveness of attribution-inspired layer ranking in comparison to a conventional magnitude-based pruning baseline. 
This section describes the experimental environment, the set of investigated architectures, and the evaluation protocol used throughout our study.

\subsection{Experimental Setup}
All experiments were performed in a standardized and reproducible environment to ensure fair comparison across models and pruning strategies. 
We used a cloud-based Google Colab instance equipped with an NVIDIA Tesla T4 GPU with 16\,GB of VRAM. 
All models were implemented using the PyTorch deep learning framework. 
For standard detection architectures, we relied on the \texttt{torchvision} library, while the YOLOv8 model was implemented using the \texttt{ultralytics} framework.

For benchmarking, all models were evaluated on the Microsoft COCO 2017 validation set. 
This dataset provides a challenging and widely adopted benchmark for object detection, enabling direct comparison with prior work. 
To compute attribution-inspired importance scores, we used a small but representative subset of the same validation set. 
Following common practice in pruning literature, the validation data was used both for attribution score computation and for evaluation; no test labels were accessed during the pruning process.

\subsection{Investigated Architectures}
To examine the generality of the proposed pruning framework, we evaluated it across a diverse set of CNN-based object detectors spanning multiple architectural paradigms. 
Specifically, we considered standard CNN backbones, including ResNet-50, MobileNetV2, and ShuffleNetV2, each equipped with an SSD-style detection head. 
We also evaluated FPN-based detectors, namely Faster R-CNN and RetinaNet, both using a ResNet-50 backbone with a Feature Pyramid Network to support multi-scale detection. 
In addition, we included a modern hybrid detector, YOLOv8n, which integrates multiple architectural innovations for efficient detection. 
Finally, to assess framework-agnostic applicability, we evaluated a lightweight detector, TinySSD, implemented in JAX/Flax. 
This diverse selection allows us to analyze the behavior of attribution-inspired layer pruning across a broad spectrum of object detection models.

\subsection{Evaluation Protocol}
A consistent evaluation protocol was adopted to enable direct comparison between unpruned baseline models, L1-norm-pruned models, and attribution-inspired pruned models. 
For each architecture, a global pruning rate of 5\% was applied, corresponding to deactivation of the bottom-ranked layers according to the respective importance metric. 
No post-pruning fine-tuning was performed, allowing us to isolate the immediate effect of the pruning criteria on model accuracy and inference efficiency.

Detection performance was measured using standard mean Average Precision (mAP) metrics. 
We report mAP at an Intersection over Union (IoU) threshold of 0.50 (\texttt{mAP@0.50}) as well as the official COCO metric averaged over IoU thresholds from 0.50 to 0.95 in increments of 0.05 (\texttt{mAP@[0.50:0.95]}). 
Inference speed was evaluated in terms of Frames Per Second (FPS) on the specified GPU hardware. 
FPS measurements were obtained by averaging runtime over 100 forward passes on fixed-size input tensors, following an initial warm-up phase to stabilize GPU performance.

Model complexity was reported using the number of trainable parameters and the theoretical number of Giga Floating-Point Operations (GFLOPs). 
Because layer-wise pruning is implemented via weight zeroing and logical deactivation, the theoretical parameter count and FLOPs remain unchanged. 
Nevertheless, the induced sparsity can lead to empirical runtime improvements depending on hardware characteristics and runtime optimizations, and measured FPS gains should therefore be interpreted as empirical rather than theoretical speedups.

\section{Results and Analysis}
\label{sec:results}
This section presents and analyzes the quantitative results of our comparative pruning experiments. 
We focus on understanding how attribution-inspired layer ranking differs from magnitude-based pruning and how these differences translate into accuracy--efficiency trade-offs across object detection architectures.

The aggregated results for a representative subset of the evaluated models are summarized in Table~\ref{tab:results}. 
The table reports detection accuracy measured by mean Average Precision (mAP) and inference speed measured in Frames Per Second (FPS), along with the relative changes compared to the unpruned baseline. 
Rather than emphasizing absolute improvements, our analysis highlights the differing behaviors induced by the two pruning criteria.

\begin{table}[ht!]
    \centering
    \caption{Aggregated performance metrics for baseline, L1-norm-pruned, and attribution-inspired pruned models across selected architectures. For each architecture, the pruned variant exhibiting the most favorable accuracy--efficiency trade-off is highlighted in bold.}
    \label{tab:results}
    \scalebox{0.9}{
    \begin{tabular}{l l c c c c c}
        \toprule
        \textbf{Model Architecture} & \textbf{Method} & \textbf{mAP@[.50:.95]} & \textbf{mAP@.50} & \textbf{FPS} & \textbf{\% $\Delta$ mAP} & \textbf{\% $\Delta$ FPS} \\
        \midrule
        \textbf{MobileNetV2} & Baseline & 0.158 & 0.205 & 29.26 & - & - \\
        & L1-Pruned & 0.157 & 0.206 & \textbf{62.95} & -0.6\% & \textbf{+115.1\%} \\
        & Attr.-Pruned & 0.156 & 0.205 & 60.60 & -1.3\% & +107.1\% \\
        \midrule
        \textbf{ResNet-50} & Baseline & 0.152 & 0.204 & 38.39 & - & - \\
        & L1-Pruned & 0.150 & 0.201 & 58.18 & -1.3\% & +51.5\% \\
        & Attr.-Pruned & 0.147 & 0.201 & \textbf{59.24} & -3.3\% & \textbf{+54.3\%} \\
        \midrule
        \textbf{ShuffleNetV2} & Baseline & 0.153 & 0.206 & 40.70 & - & - \\
        & L1-Pruned & \textbf{0.153} & 0.210 & 35.13 & \textbf{0.0\%} & -13.7\% \\
        & Attr.-Pruned & 0.150 & 0.201 & \textbf{44.78} & -2.0\% & \textbf{+10.0\%} \\
        \midrule
        \textbf{Faster R-CNN} & Baseline & 0.152 & 0.205 & 11.78 & - & - \\
        & L1-Pruned & \textbf{0.155} & 0.205 & \textbf{12.05} & \textbf{+2.0\%} & \textbf{+2.3\%} \\
        & Attr.-Pruned & 0.148 & 0.200 & 11.71 & -2.6\% & -0.6\% \\
        \midrule
        \textbf{RetinaNet} & Baseline & 0.151 & 0.201 & 11.78 & - & - \\
        & L1-Pruned & 0.149 & 0.196 & \textbf{12.51} & -1.3\% & \textbf{+6.2\%} \\
        & Attr.-Pruned & \textbf{0.151} & 0.205 & 11.67 & \textbf{0.0\%} & -0.9\% \\
        \bottomrule
    \end{tabular}
    }
\end{table}

\begin{figure}
    \centering
    \includegraphics[scale=0.3]{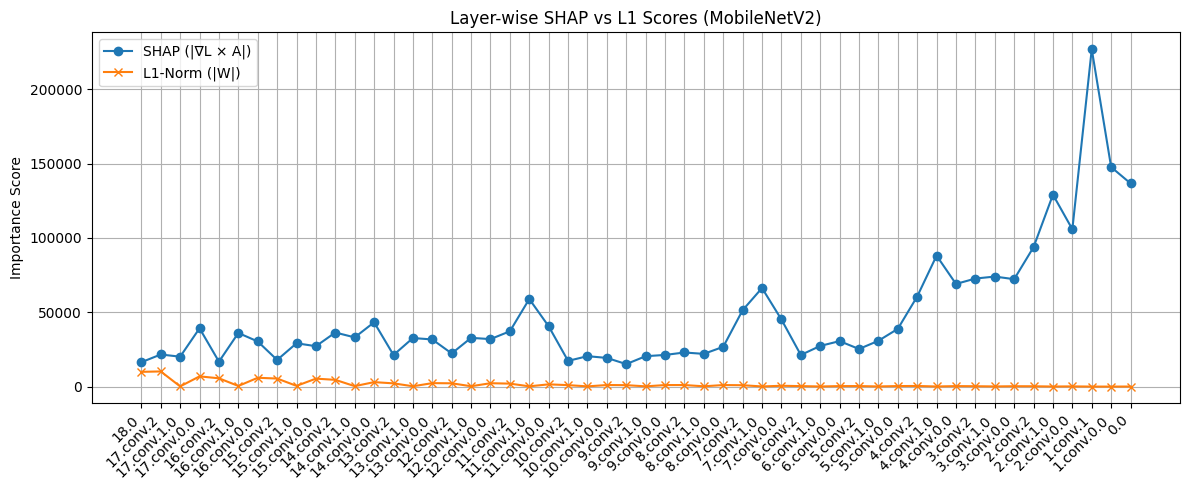}
    \caption{Comparison of layer importance rankings produced by L1-norm and attribution-inspired scoring for the MobileNetV2 architecture. The two criteria frequently disagree on which layers are assigned low importance scores.}
    \label{fig:importance}
\end{figure}

Several observations can be drawn from these results. 
For lightweight architectures that are already optimized for efficiency, such as MobileNetV2 and ShuffleNetV2, attribution-inspired pruning often identifies different pruning candidates than L1-norm ranking. 
In the case of ShuffleNetV2, attribution-inspired pruning yields an empirical increase in inference speed of 10\%, whereas L1-norm pruning leads to a notable slowdown. 
This behavior suggests that magnitude-based criteria may remove structurally small but functionally important layers in highly optimized networks, whereas attribution-inspired scores better reflect task-dependent relevance.

For more complex, FPN-based detectors such as RetinaNet, the results illustrate a clear accuracy--efficiency trade-off. 
Attribution-inspired pruning preserves the baseline detection accuracy while incurring only a marginal change in inference speed. 
In contrast, L1-norm pruning achieves modest speed improvements at the cost of reduced mAP. 
These outcomes indicate that attribution-inspired ranking can favor conservative pruning decisions that maintain predictive performance in deeper, multi-scale architectures.

A consistent pattern across architectures, illustrated in Fig.~\ref{fig:importance}, is the disagreement between magnitude-based and attribution-inspired layer rankings. 
Layers assigned low importance by L1-norm scoring are frequently different from those identified by attribution-inspired scores. 
This observation supports the premise that static weight magnitude is not always a reliable proxy for functional contribution. 
In particular, early convolutional layers, which tend to have relatively few parameters, may receive low L1-norm scores despite playing a critical role in feature extraction; attribution-inspired scoring often assigns higher importance to such layers due to their influence on the task loss.

\begin{figure}
    \centering
    \includegraphics[scale=0.6]{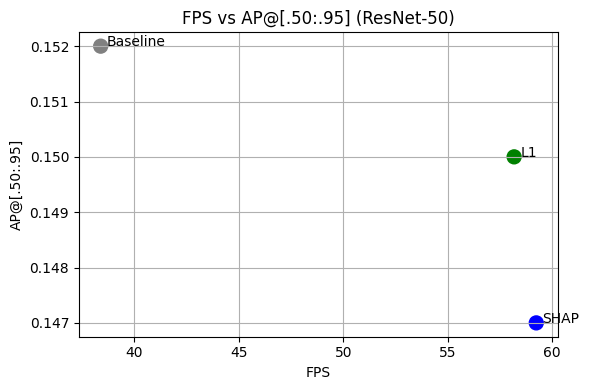}
    \caption{Accuracy--efficiency trade-off for the ResNet-50 architecture, showing mean Average Precision (mAP@[.50:.95]) versus inference speed (FPS). Different pruning criteria lead to distinct operating points.}
    \label{fig:tradeoff}
\end{figure}

These differences in layer ranking lead to distinct optimization paths, as illustrated for the ResNet-50 architecture in Fig.~\ref{fig:tradeoff}. 
Attribution-inspired pruning yields higher empirical inference speed, whereas L1-norm pruning preserves slightly higher accuracy. 
This trade-off highlights that no single pruning criterion is universally optimal and that the choice of importance metric can be guided by deployment priorities, such as prioritizing runtime efficiency or maintaining conservative accuracy guarantees.

\section{Conclusion}
\label{sec:conclusion}

This paper presented a comprehensive study of explainability-inspired, layer-wise pruning for deep neural networks in object detection tasks. 
By adopting a gradient--activation attribution inspired by SHAP and related explainability methods, we introduced a data-driven approach to estimating functional layer importance beyond static magnitude-based criteria. 
Through extensive evaluation across multiple object detection architectures, we analyzed how attribution-inspired layer ranking compares with conventional L1-norm pruning in terms of accuracy and inference efficiency.

Our experimental results reveal several consistent observations. 
First, magnitude-based and attribution-inspired importance measures frequently disagree on layer rankings, indicating that static weight magnitude is not always a reliable proxy for functional contribution. 
These differences in ranking lead to distinct pruning behaviors and accuracy--efficiency trade-offs, allowing practitioners to select pruning criteria that align with specific deployment objectives, such as prioritizing empirical inference speed or preserving conservative accuracy guarantees. 
In particular, for lightweight architectures such as ShuffleNetV2, attribution-inspired pruning avoids performance degradation observed with L1-norm pruning, while for deeper, multi-scale models such as RetinaNet, it enables accuracy-preserving pruning decisions.

The empirical analysis also highlights the importance of pruning rate selection. 
A global pruning rate of 5\%, determined through preliminary analysis, provides a practical balance between efficiency gains and accuracy preservation across architectures, whereas more aggressive pruning leads to disproportionate performance degradation. 
This observation offers actionable guidance for deploying object detection models in resource-constrained environments.

Overall, our findings suggest that explainability-inspired attribution can serve as a practical and interpretable signal for guiding layer-wise pruning in complex vision systems. 
By grounding pruning decisions in data-dependent measures of functional relevance, the proposed framework offers a principled alternative to purely magnitude-based approaches without introducing architectural modifications.

Several directions remain for future work. 
Exploring adaptive pruning rates that account for architectural characteristics, incorporating post-pruning fine-tuning while preserving interpretability, and extending the framework to newer detection models and other computer vision tasks represent promising avenues for further investigation.

% ======================
% Bibliography (LNCS)
% ======================
\bibliographystyle{splncs04}
\bibliography{ref}

% ======================
% (Optional) Appendix
% ======================
% \appendix
% \section{Appendix}

\end{document}